\pdfoutput=1

\documentclass[11pt]{article}

\usepackage[]{acl}

\usepackage{times}
\usepackage{latexsym}

\usepackage[T1]{fontenc}

\usepackage[utf8]{inputenc}
\usepackage{amsmath}
\usepackage{microtype}
\usepackage{graphicx}
\usepackage{makecell}
\usepackage[]{xcolor}

\title{Zero-Shot Aspect-Based Sentiment Analysis}

  \author{
Lei Shu$^{1}$\thanks{Work was done prior to joining Amazon.}, Hu Xu$^{2}$, Bing Liu$^{3}$, \and Jiahua Chen$^{3}$  \\ 
$^1$Amazon AWS AI\\
$^2$Facebook AI Research\\
$^3$Department of Computer Science, University of Illinois at Chicago\\
$^1$\texttt{shulindt@gmail.com}\\
$^2$\texttt{huxu@fb.com}\\
$^3$\texttt{\{liub, jiahuac2\}@uic.edu}\\ 
}

\begin{document}
\maketitle

\begin{abstract}
Aspect-based sentiment analysis (ABSA) typically requires in-domain annotated data for supervised training/fine-tuning.
It is a big challenge to scale ABSA to a large number of new domains.
This paper aims to train a unified model that performs zero-shot 
ABSA without using any annotated data for a new domain.
We propose a method called \underline{co}ntrastive post-training on \underline{r}eview \underline{N}atural Language Inference (CORN).
In this method, ABSA tasks are cast as natural language inference for zero-shot transfer. We evaluate CORN on ABSA tasks, ranging from aspect extraction (AE), aspect sentiment classification (ASC), to end-to-end aspect-based sentiment analysis (E2E ABSA), which show ABSA can be conducted without any human annotated ABSA data.
\end{abstract}

\section{Introduction}
\label{sec:introduction}

Aspect-based sentiment analysis \citep{DBLP:series/synthesis/2012Liu,DBLP:conf/nips/VaswaniSPUJGKP17,DBLP:journals/widm/ZhangWL18,xu2020dombert} is challenging because it typically requires supervised training data that differs domain-by-domain.
Given a review sentence ``The battery life of this laptop is superb,'' if we want to extract the aspect ``battery life'', one needs to have a large quantity of annotated data with terms such as ``screen'', ``keyboard'' for the \textit{laptop} domain/category. When it comes to a new domain (e.g., 
{\textit{restaurant}}
), these annotated data can hardly be re-used, so a new annotation in the restaurant domain is required.
\begin{figure}
\centering
\includegraphics[width=\columnwidth]{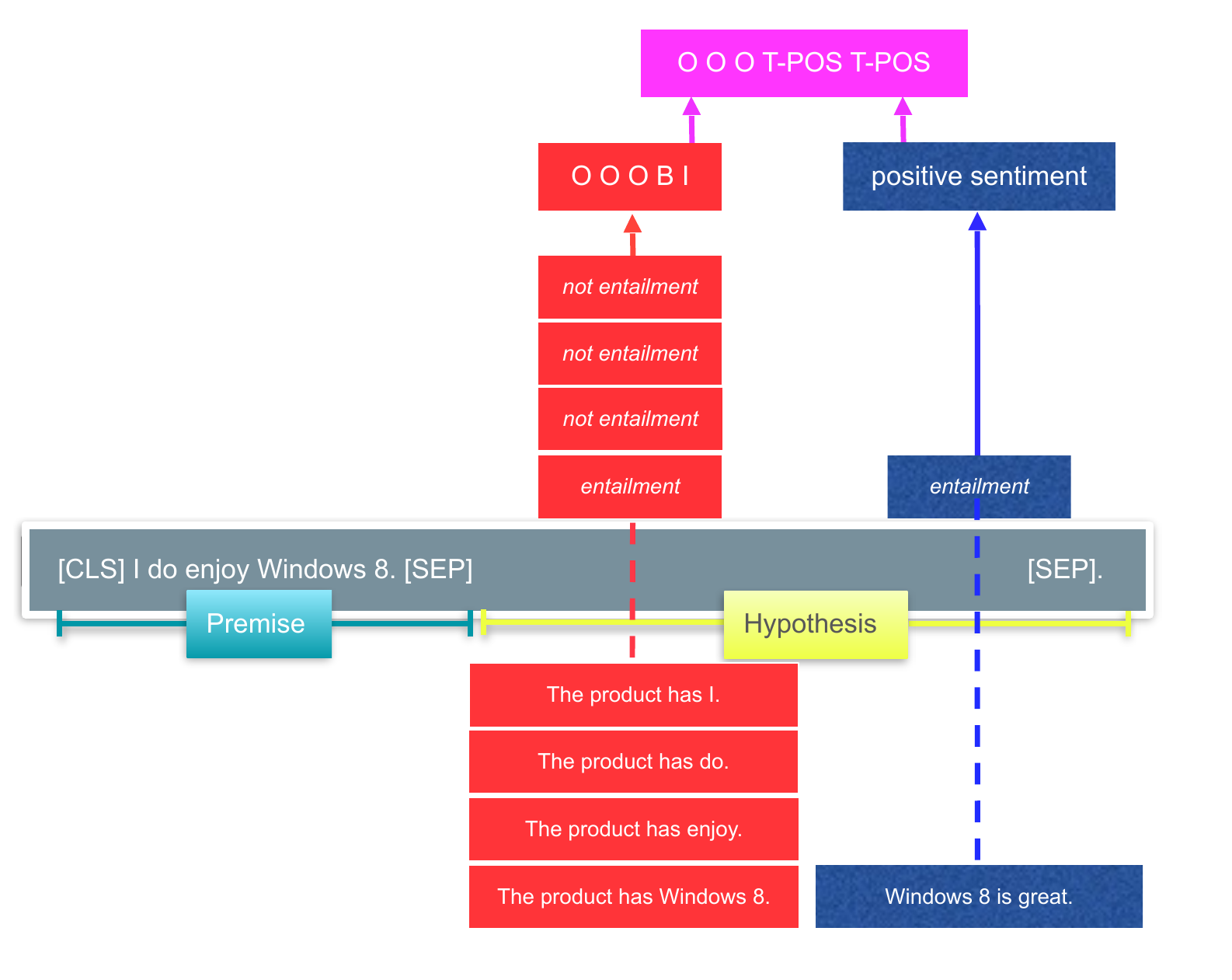}

\caption{\label{fig:model} CORN's prediction on three ABSA (AE, ASC, and E2E) tasks (Best view in color). {\color{red}\textbf{Red}}, {\color{blue}\textbf{blue}}, and {\color{magenta}\textbf{pink}} blocks indicate AE, ASC, and E2E tasks, respectively.}
\end{figure}


In this paper, we study the problem of how to learn a unified model that can be zero-shot transferred to tasks in ABSA, ranging from aspect extraction (AE) \citep{shu2017lifelong, xu2018double, shu2019controlled}, aspect sentiment classification (ASC) \citep{xu2019failure}, to end-to-end aspect-based sentiment analysis (E2E ABSA) \citep{articleaaai2019}, without requiring any human annotated ABSA data.
Although the rising of 
{pre-training/post-training\footnote{post-train takes pre-trained weights as the initialization for further training before fine-tuning using end-task annotated data.}}
on language models (LM) like BERT~\citep{DBLP:conf/naacl/DevlinCLT19} and GPT~\citep{radford2018improving}, prompt engineering ~\citep{gao2021making,emnlp/zeroasc21} has been proposed for few-shot or zero-shot transfer in sentiment classification, ABSA problems are more domain-dependent than sentiment classification. Also, certain LM losses such as those in masked language model (MLM) and next sentence prediction (NSP) in BERT\citep{DBLP:conf/naacl/DevlinCLT19} may not be suitable for zero-shot transfer to ABSA tasks.
This is because the fine-grained correlations between aspect-opinion terms and aspect-entity terms are critical, and MLM's noisy masked tokens or NSP's easy negative examples may not capture aspect-opinion and aspect-entity dependencies well.
Therefore, an ideal model must be trained on a general task and learns representations (e.g., regarding aspect-opinion terms and aspect-entity dependencies) to cover all downstream ABSA tasks and domains.

This paper proposes a method called CORN (\textit{\underline{co}ntrastive post-training on \underline{r}eview \underline{n}atural language inference}). This method casts all ABSA problems into Natural Language Inference (NLI) to infer domain-invariant relations on aspect-opinion and aspect-entity dependencies. The intuition of using NLI is to detect the logical connection in-between mentioned/inferred aspects and their correlated opinion polarities. We can create a hypothesis that includes the hypothesis aspect and hypothesis opinion polarity. If the review (premise) can entail the hypothesis, we can infer the aspect and its correlated opinion. 
{Note that the NLI task typically requires supervised training data with human annotation. 
To learn domain knowledge in reviews into CORN, we pre-process reviews into review NLI (RNLI) without human annotation.}

As the example in Figure~\ref{fig:model}, a unified model can solve all ABSA tasks. The review sentence is ``I do enjoy Windows 8''. This sentence contains the aspect ``Windows 8'' and its opinion polarity is positive. For a true AE hypothesis like ``The product has Window 8'', the NLI predicts the review sentence entails this hypothesis. For a negative hypothesis like ``The product has I.'', where ``I'' could be a token from this review sentence but not an aspect, NLI may predict that the hypothesis contradicts the review because ``I'' is not an entity associated with the product. Thus, we enumerate all possible text spans $S$ in the review like ``I'', ``do'', ``enjoy'', ``Windows 8'' to create a set of hypotheses \textit{``The product has <S>''} and then transfer their NLI predictions to AE label set \{B, I, O\}.
For ASC, we form hypotheses ``Windows 8 is great" from the span, which is predicted as an aspect like ``Windows 8'' in this example.
Likewise, the NLI predicts the review sentence entails this hypothesis for a true ASC hypothesis. And then, we transfer the NLI prediction to ASC label set \{positive, negative, neutral\} sentiment. 
Finally, we combine the prediction of AE and ASC and infer the E2E ABSA prediction.

The work makes two main contributions. (1) It proposes to perform zero-shot transfer to 3 ABSA tasks. To the best of our knowledge, zero-shot AE and E2E ABSA have not been attempted before. (2) It proposes a unified model CORN post-trained on RNLI to solve zero-shot ABSA tasks. 
Experimental results show the effectiveness of our approach without human annotation. 
\section{Related Work}







\large \textbf{Zero-shot Sentiment Analysis}. Sentiment analysis has been widely studied~\citep{hu2004mining,DBLP:journals/ftir/PangL07, DBLP:series/synthesis/2012Liu}. 
Zero-shot learning was introduced for text classification tasks using entailment as the solution~\cite{yin2019benchmarking}. Later,~\citep{yin2020universal, sainz2021label} extended entailment approaches to zero-shot and few-shot open NLP tasks and relation extraction. 
~\citet{DBLP:conf/acl/HuZGXGGCS20} focused on few-shot aspect category detection, which is a simplified few-shot aspect extraction.
~\citet{emnlp/zeroasc21} considered zero-shot and few-shot learning on ASC and DSC through fine-tuning NLI and language modeling. 
This work comprehensively solves three ABSA tasks (ASC, AE, and E2E ABSA) in a zero-shot fashion with a unified model.
Best to our knowledge, zero-shot AE and E2E ABSA, which belong to sequence labeling tasks, have not been studied before.


\large \textbf{Natural Language Inference (NLI)}. NLI considers the textual entailment problem. The problem is determining whether a t
extual hypothesis can be inferred from a premise.~\citet{DBLP:conf/mlcw/DaganGM05} introduced the task of textual entailment. Several NLI benchmarks are proposed to facilitate the research.~\citet{DBLP:conf/tac/BentivogliMDDG09} developed the 3-way decision task that considers text contradiction.~\citet{DBLP:conf/sigdial/ZhangC09} created a conversational entailment dataset that requires dialogue interpretation.~\citet{DBLP:conf/emnlp/BowmanAPM15} proposed a large NLI dataset and~\citet{DBLP:conf/naacl/WilliamsNB18} expanded the idea to 10 genres including both formal and informal corpora. These human annotation datasets incur a large labor cost. Besides, none of these works focuses on the review corpus. The approaches to solving NLI range from earlier rule-based approaches to recent deep neural network approaches~\citep{DBLP:conf/acl/MacCartneyM07,DBLP:conf/acl/ChenZLWJI17}. Several works explored the transformation of NLI tasks to other tasks. ~\citet{DBLP:conf/emnlp/ZhangHLWWYSX20} transferred the NLI knowledge on few-shot intent detection.~\citet{emnlp/zeroasc21} transferred the NLI information to zero-shot, few-shot ASC, and DSC. ~\citet{DBLP:journals/corr/abs-2104-14690} uses entailment in a few-shot scenario on 15 text classification datasets and a regression dataset.



{\large \textbf{Contrastive Learning}}. 
In our method, we adopt contrastive learning, which has been widely used in computer vision tasks~\citep{DBLP:conf/eccv/WuEY18}. Recently, contrastive learning has also been used to learn representations of input sentences in NLP~\citep{DBLP:conf/icml/RadfordKHRGASAM21}. These works use data augmentation to construct pair-wise data, which helps feature learning.~\citet{DBLP:conf/iclr/GunelDCS21} directly used the contrastive loss to make the  prediction. We use contrastive learning~\citep{DBLP:conf/nips/KhoslaTWSTIMLK20} to pull expressions with the same 
{NLI} 
labels together and push expressions with 
different 
{NLI} 
labels apart.


\section{Method}
\label{sec:method}




\begin{table}
\centering
\caption{\small Summary of notations} 
\label{table:notation}
\begin{tabular}{lp{5.5cm}}
\hline
$X$ & a review text or context\\
$I_i$ & the $i$-th input for a model\\
$A$ & an aspect \\
$S$ & a candidate span in the review \\
$Y_\textit{ASC}$, $\widehat{Y_\textit{ASC}}$ & ASC label, prediction \\
$Y_\textit{AE}$, $\widehat{Y_\textit{AE}}$ & AE labels, predictions \\
$Y_\textit{E2E}$, $\widehat{Y_\textit{E2E}}$ & E2E ABSA labels, predictions \\
$P$ & premise of an NLI example \\
$H$ & hypothesis of an NLI example \\
$Y_\textit{NLI}$, $Y_i$ & NLI label for $i$-th input $I_i$ for a model \\
\hline
\end{tabular}
\end{table}

\begin{table*}
\small
\centering
\caption{The summary of casting ABSA tasks as natural language inference (NLI). $X$ indicates a review sentence, $S$ corresponds to a (pre-chunked) span and $A$ corresponds to a given aspect. Taking the restaurant domain as examples, we use the domain label \textit{Restaurant} when constructing hypotheses. \textit{entail., neutral,  contra.} indicates entailment, neutral and contradiction in NLI. \textit{Aspect, Outside} indicates the span $S$'s label in AE; \textit{POS, NEU, NEG} indicate the given aspect $A$'s sentiment label in ASC; \textit{T-\{POS, NEU, NEG\}, Outside} indicate the span $S$'s aspect/outside labels in E2E ABSA.}
\label{table:ABSAtemplate}
\begin{tabular}{l|c|c|l|p{75mm}}
\hline
Task&Input&Premise&Hypothesis&NLI labels $\rightarrow$ ABSA labels\\\hline
AE&$(X,S)$&$X$&\textit{Restaurant} has $S$.&$\textit{entail.}\rightarrow \textit{T}, \textit{\{neutral,contra.\}}\rightarrow \textit{Outside}$\\
\hline
ASC&$(X,A)$&$X$&$A$ is great.&$\textit{entail.}\rightarrow \textit{POS}, \textit{neutral}\rightarrow \textit{NEU}, \textit{contra.}\rightarrow\textit{NEG}$\\
\hline
E2E(step1) &$(X,S)$&$X$&\textit{Restaurant} has $S$.&$\textit{entail.}\rightarrow \textit{<go to step2>}, \{\textit{neutral,contra.}\}\rightarrow \textit{Outside}$\\
E2E(step2) &$(X,S)$&$X$&$S$ is great.&$\textit{entail.}\rightarrow \textit{T-POS}, \textit{neutral}\rightarrow \textit{T-NEU}, \textit{contra.}\rightarrow \textit{T-NEG}$\\
\hline
\end{tabular}
\end{table*}

\large \textbf{Overview}. 
As discussed in the introduction section, the critical point of allowing one model to perform multiple tasks in ABSA without supervised learning is to formulate a general learning task {to post-train on}.
We adopt natural language inference (NLI) as the general task to capture subtle differences in pairs of texts.
This section consists of three parts. In Section~\ref{subsec:ABSAtemplate}, we first revisit the NLI problem. Then, we describe how three (3) ABSA tasks can be cast into three forms of NLI problems via prompting. Table~\ref{table:ABSAtemplate} summarizes our problem transformation. 
In Section~\ref{subsec:OurMNLIDataset}, we describe how we process reviews into the NLI form with pseudo labels from metadata and reviews.
We name our curated review NLI data as RNLI. Lastly, in Section~\ref{subsec:Ourtraining}, we introduce the details of post-training on RNLI following the transformation in Section~\ref{subsec:ABSAtemplate}. Table~\ref{table:notation} summarizes a (non-exhaustive) list of notations, used repeatedly in the subsequent sections.

\subsection{ABSA to NLI}
\label{subsec:ABSAtemplate}

\textbf{Natural Language Inference (NLI)}. NLI is a core NLP problem. Specifically, a sentence, a document, or paragraph might be given as the premise $P$; the task then is to infer whether a given hypothesis $H$ is implied by (\textit{entailment}), irrelevant to (\textit{neutral}) or contradicted (\textit{contradiction}) by the premise. The three relations are labeled with $Y_\textit{NLI}$ which takes one of the labels in \{\textit{entailment}, \textit{neutral}, \textit{contradiction}\}.

\textbf{Intuition of ABSA to NLI}. 
As described in Section~\ref{sec:introduction}, aspect-opinion and aspect-entity dependencies are the core problems for all ABSA tasks. 
Specifically, ASC aims to find the opinion polarity of a given aspect. AE and E2E ABSA aim to find aspects and their associated opinion polarities. We observe that when a hypothesis expresses a true aspect-opinion or aspect-entity dependency in the review, an NLI model may predict that the review entails this hypothesis. 

ASC is a text classification task.
\citet{DBLP:conf/acl/pretrainfewshot2021} designed a simple yet effective prompt/template  \textit{<aspect> is <label>} to cast the aspect and the label into a natural language sentence.
From the NLI perspective, the prompt is the hypothesis $H$, and the input text/review is the premise $P$. As such, by inferencing multiple \textit{<label>}s of different sentiment polarities, NLI can perform aspect-based sentiment classification.
Further,
AE and E2E are sequence labeling problems, which is more challenging to formulate as an NLI problem given NLI is a 3-class classification task. 
The idea is to propose multiple candidates of spans (either aspect, e.g.,  ``Window 8'' or other chunks of words, e.g., ``I''). Then, NLI can find aspects that entail aspect-opinion dependency (e.g., finding Window 8 is an aspect).
As such, we chunk words into candidates of spans and create hypotheses for every candidate. Then, we summarize the predictions from all candidate spans to get the final AE/E2E ABSA prediction. Table ~\ref{table:ABSAtemplate} is a summary of how we cast ABSA tasks as NLI. The details of the transformations are as follows.


\noindent \large \textbf{Aspect Sentiment Classification (ASC)}. 
ASC aims to classify sentiment polarity $Y_\textit{ASC}$ associated with a known aspect $A$ in an input review sentence $X$. The sentiment polarity can be one of \{\textit{positive}, \textit{neutral}, \textit{negative}\} (we abbreviate the label set as \{\textit{POS}, \textit{NEU}, \textit{NEG}\}). 
To cast this problem into an NLI problem, we take the review $X$ as premise $P$ and take prompt \textit{<aspect> is great} as hypothesis $H$. 
If the hypothesis prompt can be implied by the premise (\textit{entailment}), ASC prediction is \textit{POS}. Similarly, \textit{contradiction} of NLI is mapped to \textit{NEG} and \textit{neutral} is mapped to \textit{neutral}, respectively.

\noindent \large \textbf{Aspect Extraction (AE)}. Aspect extraction (AE) aims to find aspects that may have associated opinions in the review sentence $X$. It is typically modeled as a sequence labeling task, where each token of the review sentence is labeled as one of \{\textit{Begin}, \textit{Inside}, \textit{Outside}\} (or \{\textit{B}, \textit{I}, \textit{O}\} for brevity). Spans that are labeled as one \textit{B} and followed by zero or more \textit{I} are treated as an aspect. 
We cast AE into multiple entailment problems between the input review sentence and spans. Specifically, we take the review sentence $X$ as the premise $P$ of NLI. We first produce multiple candidates of spans via choosing chunked tokens of length within $6$ ~\cite{cui-etal-2021-template}, where each candidate $S$ will later be included in a prompt to compose one hypothesis $H$. 
Note that we craft different prompts for different domains. 
For example, for the restaurant domain, we use \textit{Restaurant has $S$} as the template prompt. 
During prediction, we simplify \textit{B} and \textit{I} as a single label \textit{T}. If the hypothesis prompt can be implied by the premise (\textit{entailment}), the AE prediction of the span is mapped to \textit{T}. Otherwise, the prediction is \textit{O}. We predict all possible $(X, S)$ pairs and summarize them to predictions of all tokens in the review $\widehat{Y_\textit{AE}}$. Note, if two spans have text overlapping and yet are assigned with different labels, we choose the span with the higher score as the final prediction to avoid conflicts.

\noindent \large \textbf{End-to-end ABSA (E2E)}.~\citet{li-etal-2019-exploiting} conducts end-to-end aspect-based sentiment analysis which treats the two tasks AE and ASC as one unique task. For example, given input $X$, end-to-end ABSA aims to predict whether a token is part of an aspect. If so, it further predicts the sentiment polarity associated with that aspect. Following the implementation of~\cite{DBLP:conf/naacl/XuLSY19}, each token of the review sentence can be labeled as one of \{\textit{T-POS}, \textit{T-NEU}, \textit{T-NEG}, \textit{O}\}, which is a combined label space of AE and ASC and $T$ implies the token is a part of aspect $A$. 
\
Inspired by the previous formulation for AE and ASC, we propose to perform zero-shot E2E ABSA in two steps, a combination of AE and ASC as shown in Table ~\ref{table:ABSAtemplate} E2E(step1) and E2E(step2). 
Specifically, we predict all possible $(X, S)$ pairs in the review. 
We first perform zero-shot AE by using the hypothesis \textit{``<Entity> has S''}. 
If the span $S$ is predicted as an aspect $T$, we go to E2E(step2) and use the hypothesis \textit{``S is great''}
to perform zero-shot ASC. 
The final E2E prediction is one of \{\textit{T-POS}, \textit{T-NEU}, \textit{T-NEG}\}. If during zero-shot AE, the span $S$ is predicted as \textit{O}, its final E2E prediction is \textit{O}. 

\subsection{Review NLI (RNLI)}
\label{subsec:OurMNLIDataset}
Existing NLI datasets are very labor-intensive to annotate, which is not scalable to a large number of domains. Instead, we pre-process reviews using simple yet effective rules~\citep{DBLP:journals/coling/QiuLBC11} to curate a review NLI data with pseudo labels, without prior knowledge of any domains. We name such pre-processed data as RNLI for simplicity.





An RNLI example is composed of a premise and hypothesis pair curated from reviews.
We define the sentiment clause as a language expression for an aspect and its associated opinion. The opinion can be positive, negative, and neutral, not presented (unknown).
Our hypotheses and premises are composed of sentiment clauses that express sentiment on aspects. We detail the process as follows.

\begin{table*}
\caption{\label{table:rnliexample} NLI labels for review sentence (Hypothesis), pre-chunked $A$spect, ground truth (Groud.) and polarity in NLI hypothesis (H-polar.).}
\small
\centering
\resizebox{0.98\textwidth}{!}{
\begin{tabular}{l|c|c|c|c}
\hline
Hypothesis&$A$spect & Ground. & H-polar. & NLI Label\\
\hline
Thats what makes studying a DVD such a good {\textbf{learning tool}}.&\textbf{learning tool}&\textit{POS}&\textit{POS}&\textit{entail.}\\\cline{1-1}\cline{4-5}
An indispensable {\textbf{learning tool}}.&&&\textit{NEU}&\textit{entail.}.\\\cline{1-1}\cline{4-5}
I found this to be the worst {\textbf{learning tool}} I have ever seen.&&&\textit{NEG}&\textit{contra.}\\\hline\hline
This is doubtless part of their {\textbf{marketing plan}}.&\textbf{marketing plan}&\textit{NEU}&\textit{POS}&\textit{neutral}\\\cline{1-1}\cline{4-5}
I finally figured out Roxio\'s {\textbf{Marketing plan}}.&&&\textit{NEU}&\textit{entail.}.\\\cline{1-1}\cline{4-5}
So my complaints are all about their {\textbf{marketing plan}}.&&&\textit{NEG}&\textit{neutral}\\\hline\hline
This version of Lotus works great with {\textbf{Windows 8}}.&\textbf{Windows 8}&\textit{NEG}&\textit{POS}&\textit{contra.}\\\cline{1-1}\cline{4-5}
you also need {\textbf{Windows 8}}&&&\textit{NEU}&\textit{entail.}\\\cline{1-1}\cline{4-5}
Unfortunately when I had to update my computer to {\textbf{Windows 8}}&&&\textit{NEG}&\textit{entail.}\\\hline

\end{tabular}
}
\end{table*}


    \textbf{Aspect Extraction.} 
    We extract aspects using double propagation~\citep{DBLP:journals/coling/QiuLBC11, shu2016lifelong}. 
    Note that double propagation requires seed aspects that could be domain-specific. To avoid building seed aspects domain-by-domain, we leverage the metadata from products' feature description in Amazon dataset~\citep{DBLP:conf/emnlp/NiLM19}.
    With double propagation, we can extract
    ``price'' and ``Windows 10'' from ``The price for windows 10 is great !''.
    
    \textbf{Aspect Polarity Extraction. }
    Given every category's aspect set, we extract clauses that contain an aspect in the aspect set from the review data under the same category. 
    Then we infer aspect's polarity with one of \{\textit{POS}, \textit{NEU}, \textit{NEG}\}) by whether its opinion term is contained in positive or negative lexicons\footnote{\url{http://www.cs.uic.edu/~liub/FBS/opinion-lexicon-English.rar}}. If the sentence does not contain words in the opinion lexicon, we treat the clause as \textit{NEU} polarity. 
    
    \textbf{RNLI Example Generation.} A review sentence can contain multiple aspects. Thus we create a premise $P$ in an RNLI example, which is randomly composed of 
    6 to 10 sentiment clauses so that it is similar to the raw review sentence. An entailed hypothesis $H_{ent}$ is a clause whose <aspect, polarity> is matched to that of one clause in premise $P$. A neutral hypothesis $H_{neu}$ means that its clause's <aspect, polarity> is not covered by the premise $P$. A contradicted hypothesis $H_{con}$ means that its clause's aspect is covered by the premise $P$, but the polarity is totally different. Let's have an example premise: 
    \begin{quote} 
    [C1] Great {\textbf{learning tool}} for children who are falling behind in math tool. [C2] I chose to give {\textbf{marketing plan}} pro 11 a shot. [C3] I am still struggling with {\textbf{Windows 8}}.
\end{quote}
    More examples of mapping from hypothesis to NLI label is shown in Table~\ref{table:rnliexample}.
    In the example premise, \textbf{learning tool} is an aspect with positive polarity. When the hypotheses mentioned the aspect and the positive opinion like ``a good learning tool'', the NLI label is entailment. If the hypotheses mention the aspect but with no opinion term like ``it is a learning tool'', the NLI label is still entailment. The reason is that a premise consisting of an aspect with a positive or negative opinion can infer a hypothesis that only mentions the aspect without opinion. However, the premise that only consists of an aspect without any opinion mentioned cannot entail the hypotheses consisting of positive or negative opinions.

\subsection{Post-training}
\label{subsec:Ourtraining}
\textbf{Transformer Encoder}. We adopt BART encoder~\citep{DBLP:conf/acl/LewisLGGMLSZ20} as the pre-trained model and post-train it on RNLI. 
For each RNLI input ($H_i$,$P_i$), we format the input as $I_i$=[CLS]+$P_i$+[SEP]+$H_i$+[SEP]. We feed the input $X_i$ into the encoding model as follows: 

\begin{align}
z_i &= \text{BART}(I_i).
\end{align}

\textbf{Supervised Contrastive Learning (SCL)}. Supervised contrastive learning on class labels ~\citet{DBLP:conf/nips/KhoslaTWSTIMLK20} trains the model to pull samples in the same class together and meanwhile push samples with different class labels away. We use this form of contrastive learning to pull expressions with the same entailment orientation closer and push expressions with different entailment orientations apart. The loss function of SCL $\mathcal{L}_\textit{SCL}$ is:

\begin{align}
    \label{eq:scl}
    \mathcal{L}_{\textit{SCL},i} &= \frac{1}{|A_i|} \sum_{j\in A_i}\log \frac{\exp (\frac{z_i\cdot z_j}{\tau})}{\sum^{N}_{k=1,i\neq k}\exp(\frac{z_i\cdot z_k}{\tau})}\\ 
    \mathcal{L}_\textit{SCL} &= \frac{1}{N} \sum_{i = 1}^{N} \mathcal{L}_{SCL,i}.
\end{align}

where $N$ is the number of examples, {$\tau$ is the temperature}, $A_i = \{j\in \{1...N\}|i\neq j, Y_i=Y_j\}$ .  


\section{Experiments}
\label{sec:experiment}
\begin{table}
\small
\centering
\caption{\small Testing Dataset Statistics.}
\label{table:testdatastatistics}
\begin{tabular}{l|l|l}
\hline
Task & Domain & Number of Examples \\ 
\hline
AE&Restaurant&\textit{T}: 892 \textit{O}: 9204 \\\cline{2-3}
&Laptop&\textit{T}: 1115 \textit{O}: 10769\\\hline
ASC&Restaurant&\textit{POS}: 728, \textit{NEU}: 196, \textit{NEG}: 196 \\\cline{2-3}
&Laptop&\textit{POS}: 341, \textit{NEU}: 128, \textit{NEG}: 169\\\hline
E2E &Restaurant&\textit{T-POS}: 1064, \textit{T-NEU}: 310,\\
&&\textit{T-NEG}: 249, \textit{O}: 10863\\\cline{2-3}
&Laptop&\textit{T-POS}: 499, \textit{T-NEU}: 330,\\
&&\textit{T-NEG}: 219, \textit{O}: 10631\\\hline
\end{tabular}
\label{tab:plain}
\end{table}

\begin{table*}
\small
\centering
\caption{\small Comparison of CORN with other baselines for zero-shot transfer to 3 ABSA tasks. Rest., Lap. corresponds to restaurant and laptop domains, respectively. $^\dagger$ are results from ~\citet{emnlp/zeroasc21}.}
\label{table:baseline}
\resizebox{0.98\textwidth}{!}{
\begin{tabular}{l|cc|cc|cc}
\hline
 Model & \multicolumn{2}{c|}{ASC (Accuracy/Macro-F$_1$)} & \multicolumn{2}{c|}{AE (Accuracy)}& \multicolumn{2}{c}{E2E (Macro-F$_1$)}\\
&Rest.&Lap.&Rest.&Lap.&Rest.&Lap.\\\hline
\multicolumn{7}{l}{\textit{Pre-training only}}\\
\hline
BERT&56.9/45.5&57.9/50.7&44.9&46.6&10.7&25.3\\\hline
Roberta&59.3/49.1&58.8/54.4&43.1&44.5&18.5&27.6\\\hline
GPT-2&\textbf{71.4}$^\dagger$/45.5$^\dagger$&60.5$^\dagger$/39.6$^\dagger$&-&-&-&- \\\hline
\multicolumn{7}{l}{\textit{Post-training}}\\
\hline
BERT+PT&61.0/49.2&60.6/53.4&45.8&48.2&27.3&31.4\\\hline
BERT+ITPT&60.3/49.1&60.2/52.4&45.3&48.6&27.5&32.6\\\hline
\multicolumn{7}{l}{\textit{Post-training on MNLI}}\\
\hline
BERT$_{\text{base}}$+MNLI&61.8$^\dagger$/57.9$^\dagger$ &58.9$^\dagger$/54.9$^\dagger$&-&-&-&- \\\hline
BART+MNLI&67.5$\pm$0.5/69.3$\pm$0.3&70.5$\pm$0.3/70.9$\pm$0.4&56.7$\pm$0.5&59.6$\pm$0.3&32.5$\pm$0.8&36.1$\pm$0.6\\\hline
BART+SCL+MNLI&68.8$\pm$0.5/69.2$\pm$0.4&\textbf{71.3$\pm$0.3}/70.3$\pm$0.2&56.9$\pm$0.3&60.0$\pm$0.3&33.9$\pm$0.5&36.8$\pm$0.4\\\hline
\multicolumn{7}{l}{\textit{Post-training on RNLI}}\\
\hline
BART+CE+RNLI&67.0$\pm$0.4/69.7$\pm$0.7&70.5$\pm$0.4/70.3$\pm$0.4&57.6$\pm$0.7&60.7$\pm$0.5&35.4$\pm$0.6&38.9$\pm$0.5\\\hline
CORN&69.7$\pm$0.4/\textbf{70.0$\pm$0.5}&70.9$\pm$0.5/\textbf{71.0$\pm$0.8}&\textbf{58.0$\pm$0.6}&\textbf{61.5$\pm$0.7}&\textbf{37.2$\pm$0.5}&\textbf{40.3$\pm$0.6}\\\hline
\end{tabular}
}

\small
\centering
\caption{\small The precision ($\mathcal{P}$) and recall ($\mathcal{R}$) of CORN and baselines.}
\label{table:baselineprecisionrecall}
\resizebox{0.98\textwidth}{!}{
\begin{tabular}{l|cc|cc|cc}
\hline
 Model & \multicolumn{2}{c|}{ASC ($\mathcal{P}$ / $\mathcal{R}$)} & \multicolumn{2}{c|}{AE ($\mathcal{P}$ / $\mathcal{R}$)}& \multicolumn{2}{c}{E2E ($\mathcal{P}$ / $\mathcal{R}$)}\\
&Rest.&Lap.&Rest.&Lap.&Rest.&Lap.\\\hline

\multicolumn{7}{l}{\textit{Pre-training only}}\\
\hline
BERT&58.6/37.4&61.9/43.2&49.2/21.9&51.6/31.6&32.7/6.4&37.1/19.2\\\hline
Roberta&57.4/43.1&67.2/45.4&47.3/29.8&59.6/34.5&29.4/13.5&40.3/21.0\\\hline
\multicolumn{7}{l}{\textit{Post-training}}\\
\hline
BERT+PT&56.4/43.8&66.1/44.9&48.9/45.1&60.6/40.7&34.7/22.5&41.7/26.7\\\hline
BERT+ITPT&56.0/44.0&65.7/43.5&50.5/44.3&59.3/42.9&31.8/24.2&44.3/30.4\\\hline
\multicolumn{7}{l}{\textit{Post-training on MNLI}}\\
\hline
BART+MNLI&71.5$\pm$0.5/66.9$\pm$0.4&\textbf{74.6$\pm$0.4}/67.1$\pm$0.4&54.3$\pm$0.6/49.3$\pm$0.5&\textbf{65.0$\pm$0.3}/44.5$\pm$0.3&35.3$\pm$0.7/30.4$\pm$1.0&43.9$\pm$0.7/30.8$\pm$0.6\\\hline
BART+SCL+MNLI&72.0$\pm$0.5/67.1$\pm$0.5&74.3$\pm$0.2/66.9$\pm$0.3&55.4$\pm$0.5/48.9$\pm$0.4&64.1$\pm$0.4/45.3$\pm$0.5&35.9$\pm$0.7/31.9$\pm$0.6&44.9$\pm$0.6/31.2$\pm$0.4\\\hline
\multicolumn{7}{l}{\textit{Post-training on RNLI}}\\
\hline
BART+CE+RNLI&72.3$\pm$0.5/\textbf{67.5$\pm$0.4}&74.1$\pm$0.4/67.3$\pm$0.5&54.9$\pm$0.7/53.8$\pm$0.5&59.7$\pm$0.6/48.2$\pm$0.4&36.3$\pm$0.6/34.5$\pm$0.5&44.5$\pm$0.4/34.0$\pm$0.5\\\hline
CORN&\textbf{73.0$\pm$0.3}/67.4$\pm$0.6&\textbf{74.6$\pm$0.6}/\textbf{67.5$\pm$0.7}&\textbf{55.6$\pm$0.5}/\textbf{54.7$\pm$0.5}&60.3$\pm$0.6/\textbf{49.7$\pm$0.8}&\textbf{38.0$\pm$0.4}/\textbf{36.5$\pm$0.7}&\textbf{45.0$\pm$0.6}/\textbf{36.8$\pm$0.5}\\\hline
\end{tabular}
}
\end{table*}

We aim to answer the following research questions in the experiment:
(1) What is the performance difference between CORN and other pre/post-trained models?
(2) Upon ablation studies of different post-train datasets and training losses, what are their respective contributions to
the whole post-training performance gain?
(3) What are the challenges and future directions of zero-shot ABSA?
We answer them in Section~\ref{subsec:result}.

\subsection{Dataset and Evaluation}
\label{subsec:dataset}
We pre-process 29 categories of data from the Amazon review dataset~\citep{DBLP:conf/emnlp/NiLM19} into RNLI data. 
To make our RNLI data generic, we generate 50k examples per category. Besides,  to avoid imbalanced distribution over product features, we take at most ten clauses, given an aspect and a sentiment polarity. The final RNLI data contains 100k examples for each label \{\textit{entailment}, \textit{neutral}, \textit{contradiction}\}, which sum up to 300k examples. In comparison, broad-coverage human-annotated MNLI dataset~\citet{DBLP:conf/naacl/WilliamsNB18} has 433k examples. 
We split 85\%, 5\% for training and validation, respectively.
 
{\large \textbf{Testing Datasets}}. To test the performance of our post-trained model, we evaluate their zero-shot transfer performance over ASC, AE, and E2E. 
Table~\ref{table:testdatastatistics} provides the statistics. 


We follow~\cite{DBLP:conf/naacl/XuLSY19} and use SemEval 2014 Task 4 for laptop domain and SemEval 2016 Task 5 for restaurant domain to evaluate zero-shot AE. We use SemEval 2014 Task 4 for laptop and restaurant reviews to evaluate zero-shot ASC. For E2E, we follow~\cite{articleaaai2019} and use their merged SemEval data to evaluate zero-shot E2E.

{\large \textbf{Metric}}. 
We report the accuracy and Macro-F1 for ASC. For AE, we compute accuracy over three AE labels. For E2E, we also compute Macro-F1 over four E2E labels. 

\subsection{Hyper-parameter Settings}
\label{subsec:label}
We take BART-large as the backbone. In the SCL loss Eq. ~\ref{eq:scl}, we set the temperature $\tau=0.1$. To accelerate the training process, we use fp16 to speed up computation. The maximum length of model pre-training is 512, and the batch size is 16. We use Adam optimizer and set the learning rate to be $1e-5$. We train 10 epochs with early termination. We report the averaged score and standard deviation of 5 random seeds.

\subsection{Compared Methods}
\label{subsec:comparemethod}
We compare our model with available zero-shot baselines. 

{\textbf{BERT$_{\text{base}}$+MNLI}}:  is proposed in ~\citep{emnlp/zeroasc21} that finetunes the BERT-base pre-trained model on MNLI datasets for zero-shot ASC. \textbf{GPT-2} is a baseline compared in this work~\citep{emnlp/zeroasc21}.


In the following paper, BART refers to BART-large, BERT refers to uncased-bert-large, and Roberta refers to Roberta-large.
To enable these transformer models to predict zero-shot ABSA tasks, we use manually created prompts, which proved to be effective in~\citet{DBLP:conf/acl/pretrainfewshot2021}. 


{\textbf{BERT}}~\citep{DBLP:conf/nips/VaswaniSPUJGKP17}: this is the original uncased BERT-large model. 

{{\textbf{BERT+PT}}~\citep{DBLP:conf/naacl/XuLSY19}: this is post-trained BERT-large with amazon and yelp reviews on general tasks including masked-language modeling (MLM) and next sentence prediction (NSP).  
}

{\textbf{Roberta}}~\citep{DBLP:journals/corr/abs-1907-11692}: this is the original Roberta-large model. 

{\textbf{BERT+ITPT}}~\citep{DBLP:conf/cncl/SunQXH19}: This is post-trained BERT on Amazon20~\citep{DBLP:conf/ACL/ChenM015}, IMDB~\citep{DBLP:conf/ACL/MaasDPHNP11}, YELP~\citep{DBLP:conf/nips/ZhangZL15} and SST-2~\citep{DBLP:conf/emnlp/SocherPWCMNP13}

{\textbf{BART+MNLI}}: fine-tune BART with MNLI datasets~\citep{DBLP:conf/naacl/WilliamsNB18}. The trained model is available online\footnote{https://huggingface.co/facebook/bart-large-mnli}. Note that this baseline uses an extra output layer to fine-tune on cross-entropy loss.


{\textbf{CORN}}: our contrastive post-pretraining on RNLI. The details of CORN are in Section~\ref{sec:method}. 
To gain further insight, we ablate various components of our model. 
First, we are interested in the contribution of RNLI. Then, we compare our model with the same post-training approach using the human-annotated NLI data MNLI and denote this method as \textbf{BART+SCL+MNLI}.
Second, we are interested in the contribution of different types of losses.
We train RNLI with cross-entropy loss and denote this method as {\textbf{BART+CE+RNLI}}.

\begin{table*}
\small
\centering
\caption{\small Qualitative study, with predictions from three models over E2E ABSA. In the AE NLI prediction, we use \textit{NE} to indicate non-entailment (neutral or contradiction) in NLI label space. {\color{red}{}Red texts} indicate wrong predictions. (Best view in color.)}
\label{table:casestudy}
\resizebox{0.98\textwidth}{!}{
\begin{tabular}{c|ccc}
\hline
Model&\multicolumn{1}{c|}{BERT+PT}&\multicolumn{1}{c|}{BART+MNLI}&\multicolumn{1}{c}{CORN}\\\hline
Review Premise&\multicolumn{1}{c|}{I was given a demonstration of Windows \;8 \;\;\;\;.}&\multicolumn{1}{c|}{I was given a demonstration of Windows 8 .}&\multicolumn{1}{c|}{I was given a demonstration of Windows 8 .}\\\hline
True Label&\multicolumn{1}{c|}{\textit{O\; O \;\;\;O \;\;O \;\;\;\;\;\;\;\;\;\;O\;\;\;\;\;\;\;\;O T-NEU T-NEU O}}&\multicolumn{1}{c|}{\textit{O\; O \;\;\;O \;\;O \;\;\;\;\;\;\;\;\;\;O\;\;\;\;\;\;\;\;O T-NEU T-NEU O}}&\multicolumn{1}{c}{\textit{O\; O \;\;\;O \;\;O \;\;\;\;\;\;\;\;\;\;O\;\;\;\;\;\;\;\;O T-NEU T-NEU O}}\\\hline
AE Hypothesis&\multicolumn{3}{c}{\textit{Product has I.} ... \textit{Product has demonstration} ... \textit{Product has Windows.} ... \textit{Product has windows 8.} ...}\\\hline
AE NLI Prediction &\multicolumn{1}{c|}{\textit{NE NE \;NE \;NE \;\;\;\;\;{\color{red}Entail.}\;\;NE\;Entail. \;\;{\color{red}NE} NE}}&\multicolumn{1}{c|}{\textit{NE NE \;NE \;NE \;\;\;\;\;{\color{red}Entail.}\;NE\;\;\;Entail. \;{\color{red}NE} NE}}&\multicolumn{1}{c|}{\textit{NE NE \;NE \;NE \;\;\;{\color{red}Entail.}\;\;\;NE\;Entail. Entail. NE}}\\\hline
AE Prediction &\multicolumn{1}{c|}{\textit{O\; O \;\;\;O \;\;O \;\;\;\;\;\;\;\;\;\;{\color{red}B}\;\;\;\;\;\;\;\;O \;\;\;\;\;B\;\;\;\;\;\;\; {\color{red}O}\;\;\; O}}&\multicolumn{1}{c|}{\textit{O\; O \;\;\;O \;\;O \;\;\;\;\;\;\;\;\;\;{\color{red}B}\;\;\;\;\;\;\;\;O \;\;\;\;\;B\;\;\;\;\;\;\; {\color{red}O}\;\;\; O}}&\multicolumn{1}{c|}{\textit{O\; O \;\;\;O \;\;O \;\;\;\;\;\;\;\;\;\;{\color{red}B}\;\;\;\;\;\;\;\;\;O \;\;\;\;\;B\;\;\;\;\;\;\; I\;\;\;\;\; O}}\\\hline
ASC Hypothesis&\multicolumn{1}{c|}{\textit{demonstration is great.},  \textit{windows is great.}}&\multicolumn{1}{c|}{\textit{demonstration is great.},  \textit{windows is great.}}&\textit{demonstration is great.},  \textit{windows 8 is great.}\\\hline
ASC NLI Prediction&\multicolumn{1}{c|}{\textit{\;\;\;\;\;\;\;\;\;\;\;\;\;{\color{red}POS}\;\;\;\;\;\;\;\;\;\;\;{\color{red}POS}}}&\multicolumn{1}{c|}{\textit{\;\;\;\;\;\;\;\;\;\;\;\;\;{\color{red}POS}\;\;\;\;\;\;\;\;\;\;\;NEU}}&\multicolumn{1}{c|}{\textit{\;\;\;\;\;\;\;\;\;\;\;\;\;\;\;\;\;\;\;\;\;\;{\color{red}POS}\;\;\;\;\;\;\;\;\;\;\;\;NEU\;\;\;\;NEU}}\\\hline
E2E Prediction&\multicolumn{1}{c|}{\textit{O\; O \;\;\;O \;\;O \;\;\;\;\;\;{\color{red}T-POS}\;\;\;\;\;O {\color{red}T-POS}\;\;\;\;\;{\color{red}O}\;\;\; O}}&\multicolumn{1}{c|}{\textit{O\; O \;\;\;O \;\;O \;\;\;\;\;\;{\color{red}T-NEU}\;\;\;\;\;O T-NEU\;\;\;\;{\color{red}O}\;\;\; O}}&\multicolumn{1}{c|}{\textit{O\; O \;\;\;O \;\;O \;\;\;\;\;\;{\color{red}T-NEU}\;\;\;\;\;O\;T-NEU\;T-NEU O}}\\\hline
\end{tabular}
}
\end{table*}

\subsection{Results}
\label{subsec:result}
{\large \textbf{Comparison with baselines.}} 
Tables~\ref{table:baseline} and~\ref{table:baselineprecisionrecall} report scores of all models' zero-shot transfer performance for the three ABSA tasks. We observe the following: (1) BART+MNLI turns out to be the strongest baseline. This shows the importance of finetuning on the NLI dataset for ABSA tasks. (2) Post-trained models achieve higher performances for all three ABSA tasks than pre-trained models, especially on E2E tasks, by 6.3\% at least (BERT+PT outperforms Roberta). This suggests that domain adaptation is essential for ABSA tasks. (3) Overall, our model performs the best, especially on AE and E2E tasks.

BART+MNLI outperforms BERT, Roberta, and post-trained BERTs on all three tasks. 
The MNLI dataset contains human-annotated examples for sentiment analysis and possession. 
Thus we use them to infer aspect-opinion and aspect-entity relations mentioned in the premise. 
However, MLM and NSP losses-based pre-trained and post-trained methods are self-supervised and cannot reflect the aspect-opinion and aspect-entity dependency well. 
BART+MNLI mostly performs better than GPT-2 and BERT$_\text{base}$+MNLI ~\citep{emnlp/zeroasc21} on ASC. It may be due to the pre-trained model configurations, prompts, and post-processing methods on NLI predictions.

In the second observation, we compare post-trained BERTs (on domain corpus) to pre-trained methods, BERT and Roberta. Their training losses are the same. Nevertheless, BERT+PT and BERT+ITPT perform slightly better than BERT and Roberta because the domain corpus contains a higher density of sentiment words and aspect words than the corpus used in BERT and Roberta.

Last but not least, CORN outperforms the baselines. In ASC, we observe that CORN performs close to the models finetuned on MNLI( BART+MNLI and BART+SCL+MNLI) in all metrics (accuracy, macro-F$_1$, precision, and recall). Considering that MNLI is human annotated while RNLI's sentiment analysis examples are curated from rules, we think that rule-based sentiment-related NLI examples can achieve a similar effect as human-annotated examples. CORN performs slightly better in the AE task than MNLI-based models on accuracy. Diving deep into Table ~\ref{table:baselineprecisionrecall}, we find that MNLI-based models have higher precision but lower recall than CORN. This suggests that RNLI can enrich the aspect terms; however, the precision of possession cannot achieve a human-annotated level as MNLI. E2E ABSA is an aggregation task over AE and ASC. CORN performs better than MNLI-based models because of its high recall of AE.


{\large \textbf{Ablation study.}} To evaluate how contrastive post-training and RNLI contribute respectively, firstly, we compare CORN to BART+SCL+MNLI. From Table~\ref{table:baseline}, CORN has performance gains on AE and E2E. This indicates that RNLI feeds the model more domain-specific information for AE and E2E ABSA tasks.
We validate our post-training approach by comparing it with BART+CE+RNLI. CORN outperforms BART+CE+RNLI on all three tasks. This shows the effectiveness of contrastive post-training.


{\large \textbf{\textbf{Qualitative Analysis}}}
We provide a case study to show the effectiveness of our model by comparing it with two baseline models, BART+MNLI and BERT+PT. Table~\ref{table:casestudy} shows a sample in the E2E task. There are two steps: (1) identifying ``Windows 8'' as an aspect (2) determining the sentiment polarity as neutral. BART+MNLI fails to extract ``8'' due to its lack of domain-specific information. The E2E prediction of BERT+PT is the worst. It fails to classify the sentiment polarity of ``Windows 8'' correctly, probably because it lacks knowledge about aspect-opinion dependencies. CORN performs the best. However, its overall performance cannot achieve supervised models.
There are limitations on the prompts used in testing and the rules used in RNLI curation. Furthermore, there exists a gap between them. As shown in Table ~\ref{table:ABSAtemplate}, the prompt for AE is \textit{``<Entity> has S''}. It indicates a possession relation. However, not all possession samples contribute to AE. For example, the ``demonstration'' in the case study ``I was given a demonstration of Windows 8.'' has the possession relation with the entity while it is not an aspect. Another premise example is that ``It does not come with the keyboard''. The ``keyboard'' is an aspect. However, the NLI prediction on the hypotheses ``The product has keyboard''  is a contradiction and indicates ``keyboard'' is not an aspect. This example suggests that the possession relation cannot cover all AE.
Nevertheless, the weaknesses aforementioned are the future directions.
 
\section{Conclusion}

This paper proposed contrastive post-training on review NLI. This method casts three ABSA tasks (AE, ASC, E2E ABSA) into natural language inference (NLI) to allow zero-shot transfer. We pre-process reviews into NLI form based on rules. As a result,  our method alleviates the need for human annotation for both ABSA and NLI. Furthermore, experimental results show that our approach achieves promising results. 
\bibliography{custom}

\begin{thebibliography}{46}
\expandafter\ifx\csname natexlab\endcsname\relax\def\natexlab#1{#1}\fi

\bibitem[{Bentivogli et~al.(2009)Bentivogli, Magnini, Dagan, Dang, and
  Giampiccolo}]{DBLP:conf/tac/BentivogliMDDG09}
Luisa Bentivogli, Bernardo Magnini, Ido Dagan, Hoa~Trang Dang, and Danilo
  Giampiccolo. 2009.
\newblock The fifth {PASCAL} recognizing textual entailment challenge.
\newblock In \emph{Proceedings of the Second Text Analysis Conference, {TAC}
  2009, Gaithersburg, Maryland, USA, November 16-17, 2009}. {NIST}.

\bibitem[{Bowman et~al.(2015)Bowman, Angeli, Potts, and
  Manning}]{DBLP:conf/emnlp/BowmanAPM15}
Samuel~R. Bowman, Gabor Angeli, Christopher Potts, and Christopher~D. Manning.
  2015.
\newblock A large annotated corpus for learning natural language inference.
\newblock In \emph{Proceedings of the 2015 Conference on Empirical Methods in
  Natural Language Processing, {EMNLP} 2015, Lisbon, Portugal, September 17-21,
  2015}, pages 632--642. The Association for Computational Linguistics.

\bibitem[{Chen et~al.(2017)Chen, Zhu, Ling, Wei, Jiang, and
  Inkpen}]{DBLP:conf/acl/ChenZLWJI17}
Qian Chen, Xiaodan Zhu, Zhen{-}Hua Ling, Si~Wei, Hui Jiang, and Diana Inkpen.
  2017.
\newblock Enhanced {LSTM} for natural language inference.
\newblock In \emph{Proceedings of the 55th Annual Meeting of the Association
  for Computational Linguistics, {ACL} 2017, Vancouver, Canada, July 30 -
  August 4, Volume 1: Long Papers}, pages 1657--1668. Association for
  Computational Linguistics.

\bibitem[{Chen et~al.(2015)Chen, Ma, and Liu}]{DBLP:conf/ACL/ChenM015}
Zhiyuan Chen, Nianzu Ma, and Bing Liu. 2015.
\newblock Lifelong learning for sentiment classification.
\newblock In \emph{Proceedings of the 53rd Annual Meeting of the Association
  for Computational Linguistics and the 7th International Joint Conference on
  Natural Language Processing of the Asian Federation of Natural Language
  Processing}, pages 750--756. The Association for Computer Linguistics.

\bibitem[{Cui et~al.(2021)Cui, Wu, Liu, Yang, and
  Zhang}]{cui-etal-2021-template}
Leyang Cui, Yu~Wu, Jian Liu, Sen Yang, and Yue Zhang. 2021.
\newblock Template-based named entity recognition using {BART}.
\newblock In \emph{Findings of the Association for Computational Linguistics:
  ACL-IJCNLP 2021}, pages 1835--1845.

\bibitem[{Dagan et~al.(2005)Dagan, Glickman, and
  Magnini}]{DBLP:conf/mlcw/DaganGM05}
Ido Dagan, Oren Glickman, and Bernardo Magnini. 2005.
\newblock The {PASCAL} recognising textual entailment challenge.
\newblock In \emph{Machine Learning Challenges, Evaluating Predictive
  Uncertainty, Visual Object Classification and Recognizing Textual Entailment,
  First {PASCAL} Machine Learning Challenges Workshop, {MLCW} 2005,
  Southampton, UK, April 11-13, 2005, Revised Selected Papers}, volume 3944 of
  \emph{Lecture Notes in Computer Science}, pages 177--190. Springer.

\bibitem[{Devlin et~al.(2019)Devlin, Chang, Lee, and
  Toutanova}]{DBLP:conf/naacl/DevlinCLT19}
Jacob Devlin, Ming{-}Wei Chang, Kenton Lee, and Kristina Toutanova. 2019.
\newblock {BERT:} pre-training of deep bidirectional transformers for language
  understanding.
\newblock In \emph{Proceedings of the 2019 Conference of the North American
  Chapter of the Association for Computational Linguistics: Human Language
  Technologies, {NAACL-HLT} 2019, Minneapolis, MN, USA, June 2-7, 2019, Volume
  1 (Long and Short Papers)}, pages 4171--4186. Association for Computational
  Linguistics.

\bibitem[{Gao et~al.(2021{\natexlab{a}})Gao, Fisch, and Chen}]{gao2021making}
Tianyu Gao, Adam Fisch, and Danqi Chen. 2021{\natexlab{a}}.
\newblock Making pre-trained language models better few-shot learners.
\newblock In \emph{Association for Computational Linguistics (ACL)}.

\bibitem[{Gao et~al.(2021{\natexlab{b}})Gao, Fisch, and
  Chen}]{DBLP:conf/acl/pretrainfewshot2021}
Tianyu Gao, Adam Fisch, and Danqi Chen. 2021{\natexlab{b}}.
\newblock Making pre-trained language models better few-shot learners.
\newblock In \emph{Proceedings of the 59th Annual Meeting of the Association
  for Computational Linguistics and the 11th International Joint Conference on
  Natural Language Processing, {ACL/IJCNLP} 2021, (Volume 1: Long Papers),
  Virtual Event, August 1-6, 2021}, pages 3816--3830. Association for
  Computational Linguistics.

\bibitem[{Gunel et~al.(2021)Gunel, Du, Conneau, and
  Stoyanov}]{DBLP:conf/iclr/GunelDCS21}
Beliz Gunel, Jingfei Du, Alexis Conneau, and Veselin Stoyanov. 2021.
\newblock Supervised contrastive learning for pre-trained language model
  fine-tuning.
\newblock In \emph{9th International Conference on Learning Representations}.
  OpenReview.net.

\bibitem[{Hu et~al.(2021)Hu, Zhao, Guo, Xue, Gao, Gao, Cheng, and
  Su}]{DBLP:conf/acl/HuZGXGGCS20}
Mengting Hu, Shiwan Zhao, Honglei Guo, Chao Xue, Hang Gao, Tiegang Gao, Renhong
  Cheng, and Zhong Su. 2021.
\newblock Multi-label few-shot learning for aspect category detection.
\newblock In \emph{Proceedings of the 59th Annual Meeting of the Association
  for Computational Linguistics and the 11th International Joint Conference on
  Natural Language Processing, {ACL/IJCNLP} 2021, (Volume 1: Long Papers),
  Virtual Event, August 1-6, 2021}, pages 6330--6340. Association for
  Computational Linguistics.

\bibitem[{Hu and Liu(2004)}]{hu2004mining}
Minqing Hu and Bing Liu. 2004.
\newblock Mining and summarizing customer reviews.
\newblock In \emph{Proceedings of the tenth ACM SIGKDD international conference
  on Knowledge discovery and data mining}, pages 168--177.

\bibitem[{Khosla et~al.(2020)Khosla, Teterwak, Wang, Sarna, Tian, Isola,
  Maschinot, Liu, and Krishnan}]{DBLP:conf/nips/KhoslaTWSTIMLK20}
Prannay Khosla, Piotr Teterwak, Chen Wang, Aaron Sarna, Yonglong Tian, Phillip
  Isola, Aaron Maschinot, Ce~Liu, and Dilip Krishnan. 2020.
\newblock Supervised contrastive learning.
\newblock In \emph{Advances in Neural Information Processing Systems 33: Annual
  Conference on Neural Information Processing Systems}.

\bibitem[{Lewis et~al.(2020)Lewis, Liu, Goyal, Ghazvininejad, Mohamed, Levy,
  Stoyanov, and Zettlemoyer}]{DBLP:conf/acl/LewisLGGMLSZ20}
Mike Lewis, Yinhan Liu, Naman Goyal, Marjan Ghazvininejad, Abdelrahman Mohamed,
  Omer Levy, Veselin Stoyanov, and Luke Zettlemoyer. 2020.
\newblock {BART:} denoising sequence-to-sequence pre-training for natural
  language generation, translation, and comprehension.
\newblock In \emph{Proceedings of the 58th Annual Meeting of the Association
  for Computational Linguistics, {ACL} 2020, Online, July 5-10, 2020}, pages
  7871--7880. Association for Computational Linguistics.

\bibitem[{Li et~al.(2019{\natexlab{a}})Li, Bing, Li, and Lam}]{articleaaai2019}
Xin Li, Lidong Bing, Piji Li, and Wai Lam. 2019{\natexlab{a}}.
\newblock A unified model for opinion target extraction and target sentiment
  prediction.
\newblock \emph{Proceedings of the AAAI Conference on Artificial Intelligence},
  33:6714--6721.

\bibitem[{Li et~al.(2019{\natexlab{b}})Li, Bing, Zhang, and
  Lam}]{li-etal-2019-exploiting}
Xin Li, Lidong Bing, Wenxuan Zhang, and Wai Lam. 2019{\natexlab{b}}.
\newblock \href {https://doi.org/10.18653/v1/D19-5505} {Exploiting {BERT} for
  end-to-end aspect-based sentiment analysis}.
\newblock In \emph{Proceedings of the 5th Workshop on Noisy User-generated Text
  (W-NUT 2019)}, pages 34--41, Hong Kong, China. Association for Computational
  Linguistics.

\bibitem[{Liu(2012)}]{DBLP:series/synthesis/2012Liu}
Bing Liu. 2012.
\newblock \emph{Sentiment Analysis and Opinion Mining}.
\newblock Synthesis Lectures on Human Language Technologies. Morgan {\&}
  Claypool Publishers.

\bibitem[{Liu et~al.(2019)Liu, Ott, Goyal, Du, Joshi, Chen, Levy, Lewis,
  Zettlemoyer, and Stoyanov}]{DBLP:journals/corr/abs-1907-11692}
Yinhan Liu, Myle Ott, Naman Goyal, Jingfei Du, Mandar Joshi, Danqi Chen, Omer
  Levy, Mike Lewis, Luke Zettlemoyer, and Veselin Stoyanov. 2019.
\newblock Roberta: {A} robustly optimized {BERT} pretraining approach.
\newblock \emph{CoRR}, abs/1907.11692.

\bibitem[{Maas et~al.(2011)Maas, Daly, Pham, Huang, Ng, and
  Potts}]{DBLP:conf/ACL/MaasDPHNP11}
Andrew~L. Maas, Raymond~E. Daly, Peter~T. Pham, Dan Huang, Andrew~Y. Ng, and
  Christopher Potts. 2011.
\newblock Learning word vectors for sentiment analysis.
\newblock In \emph{The 49th Annual Meeting of the Association for Computational
  Linguistics: Human Language Technologies, Proceedings of the Conference},
  pages 142--150. The Association for Computer Linguistics.

\bibitem[{MacCartney and Manning(2007)}]{DBLP:conf/acl/MacCartneyM07}
Bill MacCartney and Christopher~D. Manning. 2007.
\newblock Natural logic for textual inference.
\newblock In \emph{Proceedings of the ACL-PASCAL@ACL 2007 Workshop on Textual
  Entailment and Paraphrasing, Prague, Czech Republic, June 28-29, 2007}, pages
  193--200. Association for Computational Linguistics.

\bibitem[{Ni et~al.(2019)Ni, Li, and McAuley}]{DBLP:conf/emnlp/NiLM19}
Jianmo Ni, Jiacheng Li, and Julian~J. McAuley. 2019.
\newblock Justifying recommendations using distantly-labeled reviews and
  fine-grained aspects.
\newblock In \emph{Proceedings of the 2019 Conference on Empirical Methods in
  Natural Language Processing and the 9th International Joint Conference on
  Natural Language Processing, {EMNLP-IJCNLP} 2019, Hong Kong, China, November
  3-7, 2019}, pages 188--197. Association for Computational Linguistics.

\bibitem[{Pang and Lee(2008)}]{DBLP:journals/ftir/PangL07}
Bo~Pang and Lillian Lee. 2008.
\newblock Opinion mining and sentiment analysis.
\newblock \emph{Found. Trends Inf. Retr.}, 2(1-2):1--135.

\bibitem[{Qiu et~al.(2011)Qiu, Liu, Bu, and
  Chen}]{DBLP:journals/coling/QiuLBC11}
Guang Qiu, Bing Liu, Jiajun Bu, and Chun Chen. 2011.
\newblock Opinion word expansion and target extraction through double
  propagation.
\newblock \emph{Comput. Linguistics}, 37(1):9--27.

\bibitem[{Radford et~al.(2021)Radford, Kim, Hallacy, Ramesh, Goh, Agarwal,
  Sastry, Askell, Mishkin, Clark, Krueger, and
  Sutskever}]{DBLP:conf/icml/RadfordKHRGASAM21}
Alec Radford, Jong~Wook Kim, Chris Hallacy, Aditya Ramesh, Gabriel Goh,
  Sandhini Agarwal, Girish Sastry, Amanda Askell, Pamela Mishkin, Jack Clark,
  Gretchen Krueger, and Ilya Sutskever. 2021.
\newblock Learning transferable visual models from natural language
  supervision.
\newblock In \emph{Proceedings of the 38th International Conference on Machine
  Learning}, volume 139 of \emph{Proceedings of Machine Learning Research},
  pages 8748--8763. {PMLR}.

\bibitem[{Radford et~al.(2018)Radford, Narasimhan, Salimans, and
  Sutskever}]{radford2018improving}
Alec Radford, Karthik Narasimhan, Tim Salimans, and Ilya Sutskever. 2018.
\newblock \href
  {https://cdn.openai.com/research-covers/language-unsupervised/language_understanding_paper.pdf}
  {Improving language understanding by generative pre-training}.
\newblock \emph{openai}.

\bibitem[{Sainz et~al.(2021)Sainz, de~Lacalle, Labaka, Barrena, and
  Agirre}]{sainz2021label}
Oscar Sainz, Oier~Lopez de~Lacalle, Gorka Labaka, Ander Barrena, and Eneko
  Agirre. 2021.
\newblock Label verbalization and entailment for effective zero and few-shot
  relation extraction.
\newblock In \emph{Proceedings of the 2021 Conference on Empirical Methods in
  Natural Language Processing}, pages 1199--1212.

\bibitem[{Seoh et~al.(2021)Seoh, Birle, Tak, Chang, Pinette, and
  Hough}]{emnlp/zeroasc21}
Ronald Seoh, Ian Birle, Mrinal Tak, Haw-Shiuan Chang, Brian Pinette, and Alfred
  Hough. 2021.
\newblock Open aspect target sentiment classification with natural language
  prompts.
\newblock In \emph{Proceedings of the 2021 Conference on Empirical Methods in
  Natural Language Processing , {EMNLP} 2021, Punta Cana, Dominican Republic
  November 7-11, 2021}. Association for Computational Linguistics.

\bibitem[{Shu et~al.(2016)Shu, Liu, Xu, and Kim}]{shu2016lifelong}
Lei Shu, Bing Liu, Hu~Xu, and Annice Kim. 2016.
\newblock Lifelong-rl: Lifelong relaxation labeling for separating entities and
  aspects in opinion targets.
\newblock In \emph{Proceedings of the 2016 Conference on Empirical Methods in
  Natural Language Processing}, pages 225--235.

\bibitem[{Shu et~al.(2017)Shu, Xu, and Liu}]{shu2017lifelong}
Lei Shu, Hu~Xu, and Bing Liu. 2017.
\newblock Lifelong learning crf for supervised aspect extraction.
\newblock In \emph{Proceedings of the 55th Annual Meeting of the Association
  for Computational Linguistics (Volume 2: Short Papers)}, pages 148--154.

\bibitem[{Shu et~al.(2019)Shu, Xu, and Liu}]{shu2019controlled}
Lei Shu, Hu~Xu, and Bing Liu. 2019.
\newblock Controlled cnn-based sequence labeling for aspect extraction.
\newblock \emph{arXiv preprint arXiv:1905.06407}.

\bibitem[{Socher et~al.(2013)Socher, Perelygin, Wu, Chuang, Manning, Ng, and
  Potts}]{DBLP:conf/emnlp/SocherPWCMNP13}
Richard Socher, Alex Perelygin, Jean Wu, Jason Chuang, Christopher~D. Manning,
  Andrew~Y. Ng, and Christopher Potts. 2013.
\newblock Recursive deep models for semantic compositionality over a sentiment
  treebank.
\newblock In \emph{Proceedings of the 2013 Conference on Empirical Methods in
  Natural Language Processing}, pages 1631--1642. ACL.

\bibitem[{Sun et~al.(2019)Sun, Qiu, Xu, and Huang}]{DBLP:conf/cncl/SunQXH19}
Chi Sun, Xipeng Qiu, Yige Xu, and Xuanjing Huang. 2019.
\newblock How to fine-tune {BERT} for text classification?
\newblock In \emph{Chinese Computational Linguistics - 18th China National
  Conference}, volume 11856 of \emph{Lecture Notes in Computer Science}, pages
  194--206. Springer.

\bibitem[{Vaswani et~al.(2017)Vaswani, Shazeer, Parmar, Uszkoreit, Jones,
  Gomez, Kaiser, and Polosukhin}]{DBLP:conf/nips/VaswaniSPUJGKP17}
Ashish Vaswani, Noam Shazeer, Niki Parmar, Jakob Uszkoreit, Llion Jones,
  Aidan~N. Gomez, Lukasz Kaiser, and Illia Polosukhin. 2017.
\newblock Attention is all you need.
\newblock In \emph{Advances in Neural Information Processing Systems 30: Annual
  Conference on Neural Information Processing Systems}, pages 5998--6008.

\bibitem[{Wang et~al.(2021)Wang, Fang, Khabsa, Mao, and
  Ma}]{DBLP:journals/corr/abs-2104-14690}
Sinong Wang, Han Fang, Madian Khabsa, Hanzi Mao, and Hao Ma. 2021.
\newblock Entailment as few-shot learner.
\newblock \emph{CoRR}, abs/2104.14690.

\bibitem[{Williams et~al.(2018)Williams, Nangia, and
  Bowman}]{DBLP:conf/naacl/WilliamsNB18}
Adina Williams, Nikita Nangia, and Samuel~R. Bowman. 2018.
\newblock A broad-coverage challenge corpus for sentence understanding through
  inference.
\newblock In \emph{Proceedings of the 2018 Conference of the North American
  Chapter of the Association for Computational Linguistics: Human Language
  Technologies, {NAACL-HLT} 2018, New Orleans, Louisiana, USA, June 1-6, 2018,
  Volume 1 (Long Papers)}, pages 1112--1122. Association for Computational
  Linguistics.

\bibitem[{Wu et~al.(2018)Wu, Efros, and Yu}]{DBLP:conf/eccv/WuEY18}
Zhirong Wu, Alexei~A. Efros, and Stella~X. Yu. 2018.
\newblock Improving generalization via scalable neighborhood component
  analysis.
\newblock In \emph{European Conference Computer Vision}, volume 11211 of
  \emph{Lecture Notes in Computer Science}, pages 712--728. Springer.

\bibitem[{Xu et~al.(2018)Xu, Liu, Shu, and Philip}]{xu2018double}
Hu~Xu, Bing Liu, Lei Shu, and S~Yu Philip. 2018.
\newblock Double embeddings and cnn-based sequence labeling for aspect
  extraction.
\newblock In \emph{Proceedings of the 56th Annual Meeting of the Association
  for Computational Linguistics (Volume 2: Short Papers)}, pages 592--598.

\bibitem[{Xu et~al.(2020)Xu, Liu, Shu, and Philip}]{xu2020dombert}
Hu~Xu, Bing Liu, Lei Shu, and S~Yu Philip. 2020.
\newblock Dombert: Domain-oriented language model for aspect-based sentiment
  analysis.
\newblock In \emph{Proceedings of the 2020 Conference on Empirical Methods in
  Natural Language Processing: Findings}, pages 1725--1731.

\bibitem[{Xu et~al.(2019{\natexlab{a}})Xu, Liu, Shu, and
  Yu}]{DBLP:conf/naacl/XuLSY19}
Hu~Xu, Bing Liu, Lei Shu, and Philip~S. Yu. 2019{\natexlab{a}}.
\newblock {BERT} post-training for review reading comprehension and
  aspect-based sentiment analysis.
\newblock In \emph{Proceedings of the 2019 Conference of the North American
  Chapter of the Association for Computational Linguistics: Human Language
  Technologies, {NAACL-HLT} 2019, Minneapolis, MN, USA, June 2-7, 2019, Volume
  1 (Long and Short Papers)}, pages 2324--2335. Association for Computational
  Linguistics.

\bibitem[{Xu et~al.(2019{\natexlab{b}})Xu, Liu, Shu, and Yu}]{xu2019failure}
Hu~Xu, Bing Liu, Lei Shu, and Philip~S Yu. 2019{\natexlab{b}}.
\newblock A failure of aspect sentiment classifiers and an adaptive
  re-weighting solution.
\newblock \emph{arXiv preprint arXiv:1911.01460}.

\bibitem[{Yin et~al.(2019)Yin, Hay, and Roth}]{yin2019benchmarking}
Wenpeng Yin, Jamaal Hay, and Dan Roth. 2019.
\newblock Benchmarking zero-shot text classification: Datasets, evaluation and
  entailment approach.
\newblock In \emph{Proceedings of the 2019 Conference on Empirical Methods in
  Natural Language Processing and the 9th International Joint Conference on
  Natural Language Processing (EMNLP-IJCNLP)}, pages 3914--3923.

\bibitem[{Yin et~al.(2020)Yin, Rajani, Radev, Socher, and
  Xiong}]{yin2020universal}
Wenpeng Yin, Nazneen~Fatema Rajani, Dragomir Radev, Richard Socher, and Caiming
  Xiong. 2020.
\newblock Universal natural language processing with limited annotations: Try
  few-shot textual entailment as a start.
\newblock In \emph{Proceedings of the 2020 Conference on Empirical Methods in
  Natural Language Processing (EMNLP)}, pages 8229--8239.

\bibitem[{Zhang and Chai(2009)}]{DBLP:conf/sigdial/ZhangC09}
Chen Zhang and Joyce~Yue Chai. 2009.
\newblock What do we know about conversation participants: Experiments on
  conversation entailment.
\newblock In \emph{Proceedings of the {SIGDIAL} 2009 Conference, The 10th
  Annual Meeting of the Special Interest Group on Discourse and Dialogue, 11-12
  September 2009, London, {UK}}, pages 206--215. The Association for Computer
  Linguistics.

\bibitem[{Zhang et~al.(2020)Zhang, Hashimoto, Liu, Wu, Wan, Yu, Socher, and
  Xiong}]{DBLP:conf/emnlp/ZhangHLWWYSX20}
Jianguo Zhang, Kazuma Hashimoto, Wenhao Liu, Chien{-}Sheng Wu, Yao Wan,
  Philip~S. Yu, Richard Socher, and Caiming Xiong. 2020.
\newblock Discriminative nearest neighbor few-shot intent detection by
  transferring natural language inference.
\newblock In \emph{Proceedings of the 2020 Conference on Empirical Methods in
  Natural Language Processing, {EMNLP} 2020, Online, November 16-20, 2020},
  pages 5064--5082. Association for Computational Linguistics.

\bibitem[{Zhang et~al.(2018)Zhang, Wang, and
  Liu}]{DBLP:journals/widm/ZhangWL18}
Lei Zhang, Shuai Wang, and Bing Liu. 2018.
\newblock Deep learning for sentiment analysis: {A} survey.
\newblock \emph{Wiley Interdiscip. Rev. Data Min. Knowl. Discov.}, 8(4).

\bibitem[{Zhang et~al.(2015)Zhang, Zhao, and LeCun}]{DBLP:conf/nips/ZhangZL15}
Xiang Zhang, Junbo~Jake Zhao, and Yann LeCun. 2015.
\newblock Character-level convolutional networks for text classification.
\newblock In \emph{Advances in Neural Information Processing Systems 28: Annual
  Conference on Neural Information Processing Systems}, pages 649--657.

\end{thebibliography}
\bibliographystyle{acl_natbib}



\end{document}